\documentclass[cameraready]{Interspeech}

\usepackage{multirow}
\usepackage{booktabs}
\usepackage{multirow}
\usepackage{colortbl}
\usepackage[table]{xcolor}

\title{A Conflict-Aware Penalty and Statistical Loss Framework for Balancing Modalities and Enhancing Stability in Multimodal Sentiment Analysis}

\author[affiliation={1}]{Jianheng}{Dai}
\author[affiliation={1}]{Jiazhang}{Liang}
\author[affiliation={1}, correspondingauthor]{Sijie}{Mai}

\address{
    $^1$ School of Computer Science, South China Normal University, Guangzhou, Guangdong, China
}

\email{20232131081@m.scnu.edu.cn, 20232132016@m.scnu.edu.cn, sijiemai@m.scnu.edu.cn}


\usepackage{comment}

\begin{document}

\maketitle

\begin{abstract}
Multimodal Sentiment Analysis (MSA) fuses text, acoustic, and visual streams to infer sentiment. Because pre-trained text encoders are far more expressive than their acoustic and visual counterparts, the text modality tends to dominate optimization, suppressing weaker modalities and inducing gradient norm conflicts that destabilize training. To address this, we propose a Conflict-aware Penalty (CP) that detects and penalizes gradient norm conflicts at each training step, and a Statistical Loss (SL) that aligns predicted distribution statistics with empirical input statistics. Crucially, CP prevents dominant modality gradients from interfering with the SL objective, enabling synergistic training within a unified framework incorporating adaptive modality encoding, gated cross-modal fusion, and unimodal auxiliary heads. Experiments on CMU-MOSI demonstrate state-of-the-art performance, with ablation studies confirming the effectiveness of each component.
\end{abstract}

\section{Introduction}

Multimodal Sentiment Analysis (MSA) aims to infer human sentiment by jointly
modeling text, acoustic, and visual streams~\cite{zadeh2016mosi,zadeh2018mosei}.
Pre-trained language models such as BERT~\cite{devlin2019bert} and
DeBERTa~\cite{he2021deberta} have dramatically raised the ceiling for
text-based understanding, and a wave of fusion architectures---ranging from
tensor-based methods~\cite{zadeh2017tfn,liu2018lmf} and cross-modal
transformers~\cite{tsai2019mult} to contrastive and generative
approaches~\cite{yang2023confede,wu2024hydiscgan}---have exploited these
backbones to push benchmark scores on CMU-MOSI and CMU-MOSEI to new heights.
Despite this progress, two fundamental challenges continue to limit the
reliability and generalisability of MSA models.

\textbf{Challenge 1: Modality imbalance.}
Because pre-trained text encoders are far more expressive than their acoustic
and visual counterparts, the text modality tends to dominate the optimisation
process and suppress the gradient signal of weaker encoders, a phenomenon
termed \emph{modality laziness}~\cite{huang2022competition}.
This imbalance has been observed broadly in multimodal learning: dominant
modalities converge faster and subsequently crowd out others, preventing the
model from exploiting complementary cross-modal
information~\cite{wang2020makes,peng2022ogm}.
On-the-Fly Gradient Modulation (OGM)~\cite{peng2022ogm,on_the_fly_modulation_2024}
mitigates this by scaling each modality's gradient according to its relative
training error. However, OGM operates on performance ratios alone and cannot
detect a subtler failure mode: a modality whose prediction error is already
lower than its peers may still impose a disproportionately \emph{large gradient
norm} on the shared parameters, continuing to suppress weaker modalities even
after OGM has intervened. We call this a \emph{gradient norm conflict}, and
argue that resolving it requires explicit monitoring and penalisation of
gradient magnitudes---something no prior method addresses directly.

\textbf{Challenge 2: Instability of distributional regularisation.}
Aligning the predicted latent statistics with the empirical input statistics
(mean and variance) provides a useful inductive bias that encourages encoders
to preserve the distributional structure of raw features, analogous in spirit
to moment-matching objectives in domain
adaptation~\cite{sun2016deep} and variational
inference~\cite{kingma2014vae}. In MSA, such a \emph{Statistical Loss} (SL)
can act as a strong regulariser, as our ablation confirms (row B3 in
Table~\ref{tab:ablation}). However, when combined with gradient modulation, the
dominant modality's gradient pressure interferes with the moment-matching
objective, driving the optimisation into an unstable regime and causing a
dramatic performance collapse (rows A5 and C4). This coupling between gradient
modulation and distributional regularisation has not previously been identified
or addressed in the MSA literature.

\textbf{Proposed solution.}
We introduce a \textbf{Conflict-aware Penalty (CP)} that detects gradient norm
conflicts at each training step and applies a multiplicative penalty to the
offending modality's gradient coefficient. CP is lightweight, architecture-agnostic,
and operates within a scheduled epoch window to stabilise early training while
permitting unconstrained convergence later. Crucially, CP resolves \emph{both}
challenges simultaneously: it directly suppresses gradient norm conflicts (Challenge 1)
and, by keeping gradient pressure balanced, prevents the dominant modality from
destabilising the SL objective (Challenge 2). We build CP into a unified
multimodal framework that also incorporates Adaptive Modality Encoding (AME)
with reparameterisation~\cite{kingma2014vae}, gated cross-modal
fusion~\cite{arevalo2017gated}, unimodal auxiliary heads~\cite{yu2021selfmm},
and reconstruction regularisation. The resulting model achieves
\textbf{89.31\% Acc-2} and \textbf{0.638 MAE} on CMU-MOSI, outperforming
the previous state of the art, ITHP~\cite{xiao2024neuroinspiredinformationtheoretichierarchicalperception},
by a clear margin, while also attaining the highest Pearson correlation on
CMU-MOSEI.

\noindent\textbf{Contributions.}
\begin{itemize}
  \item We identify and formalise the \textbf{gradient norm conflict} problem
        in multimodal optimisation and introduce a \textbf{Conflict-aware Penalty (CP)}
        to resolve it.
  \item We show that \textbf{Statistical Loss (SL)} is a powerful regulariser
        for MSA but collapses without CP; together, CP+SL yield stable,
        mutually reinforcing training dynamics.
  \item We integrate CP and SL into a unified MSA framework that achieves
        state-of-the-art results on CMU-MOSI and competitive results on
        CMU-MOSEI. Code and hyperparameter configurations will be released
        upon publication.
\end{itemize}

\section{Method}

\begin{figure*}[t]
  \centering
  \includegraphics[width=\linewidth]{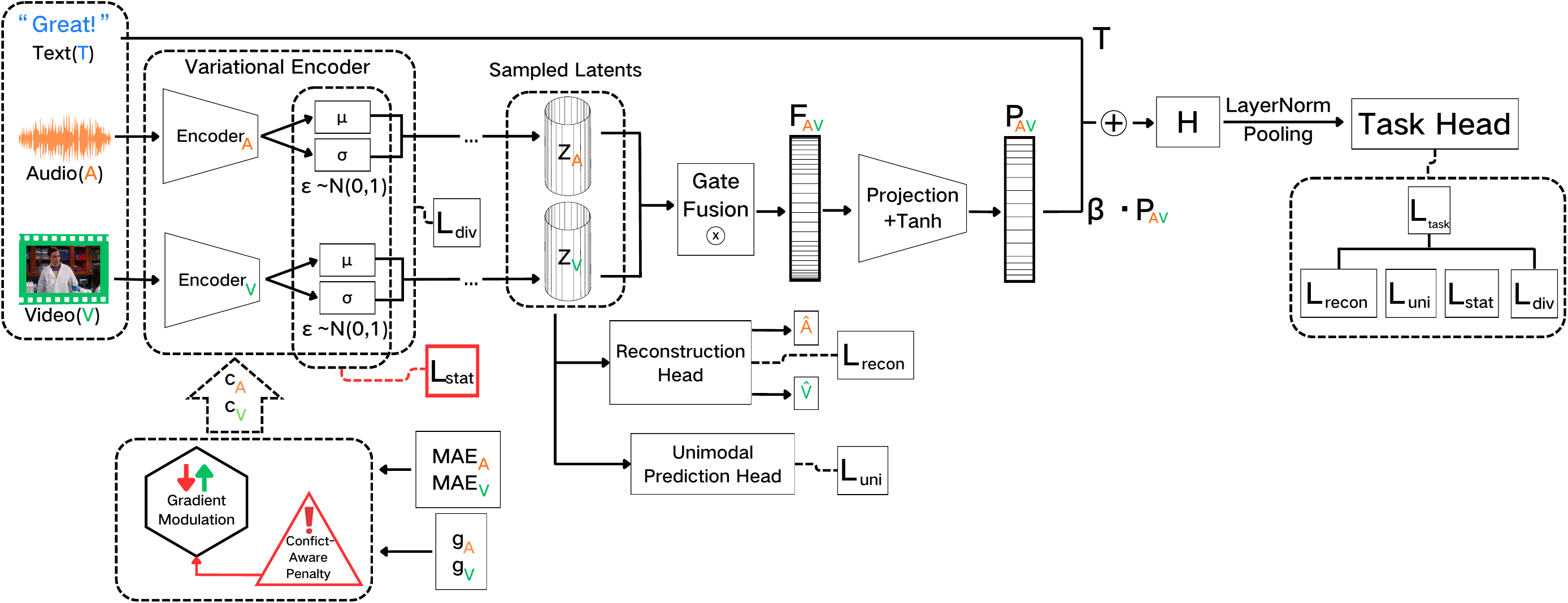}
  \caption{Overview of the Multimodal Fusion Method with Gradient Modulation and Conflict-Aware Penalty}
  \label{fig:Method}
\end{figure*}

We design a multimodal fusion model (Fig.~\ref{fig:Method}) that processes text, acoustic, and visual inputs through the following pipeline: (i) adaptive modality encoding with reparameterization, (ii) gated cross-modal fusion with residual injection into the text backbone, (iii) final prediction via a task head, and (iv) auxiliary supervision via reconstruction and unimodal heads. Our two core contributions---\textbf{Conflict-aware Penalty (CP)} and \textbf{Statistical Loss (SL)}---are inserted into steps (i) and (iv) respectively to balance modality optimization and regularize feature distributions. Given token-level text features $\mathbf{T}\in\mathbb{R}^{B\times L\times d_t}$, acoustic features $\mathbf{A}\in\mathbb{R}^{B\times L\times d_a}$, and visual features $\mathbf{V}\in\mathbb{R}^{B\times L\times d_v}$, the model produces fused representations for downstream prediction.

Step 1: Adaptive Modality Encoding (AME).
We encode each non-text modality $m\in\{a,v\}$ into a mean and variance, and sample latent features using the reparameterization trick:
\begin{align}
  \boldsymbol{\mu}_m &= f^{(\mu)}_m(\mathbf{X}_m), \nonumber \\
  \boldsymbol{\sigma}^2_m &= f^{(\sigma^2)}_m(\mathbf{X}_m), \nonumber \\
  \mathbf{Z}_m &= \boldsymbol{\mu}_m + \boldsymbol{\epsilon}\odot \sqrt{\boldsymbol{\sigma}^2_m+\varepsilon},
  \label{equation:reparam}
\end{align}
where $\mathbf{X}_a=\mathbf{A}$, $\mathbf{X}_v=\mathbf{V}$, and $\boldsymbol{\epsilon}\sim\mathcal{N}(\mathbf{0},\mathbf{I})$.

Step 2: Gated Cross-Modal Fusion and Residual Injection.
The sampled features $\mathbf{Z}_a$ and $\mathbf{Z}_v$ are fused by a configurable fusion block (e.g., gated) to obtain $\mathbf{F}_{av}\in\mathbb{R}^{B\times L\times d_f}$. We project $\mathbf{F}_{av}$ to the text hidden size and inject it through a residual shift:
\begin{align}
  \mathbf{P}_{av} &= \tanh(\mathbf{W}_p \mathbf{F}_{av}), \nonumber \\
  \mathbf{H} &= \mathbf{T} + \beta \cdot \mathbf{P}_{av},
  \label{equation:residual}
\end{align}
where $\beta$ controls the contribution of the multimodal signal. The fused sequence $\mathbf{H}$ is subsequently normalized, pooled, and fed to the task head for final prediction.

Step 3: Auxiliary Supervision and Statistical Loss (SL).
To prevent unimodal information loss, we reconstruct the original acoustic and visual inputs from the sampled latents ($\hat{\mathbf{A}},\hat{\mathbf{V}}$) and compute unimodal auxiliary predictions. The reconstruction loss $\mathcal{L}_{\mathrm{recon}}$ penalizes reconstruction errors of $\hat{\mathbf{A}}$ and $\hat{\mathbf{V}}$, and the unimodal loss $\mathcal{L}_{\mathrm{uni}}$ is computed from unimodal heads.

The \textbf{Statistical Loss (SL)} regularizes the encoder outputs by aligning their predicted distribution statistics with the empirical input statistics. Let $\hat{\mu}_m, \hat{\sigma}^2_m$ denote the empirical mean and variance of input modality $m \in \{a,v\}$ computed along the feature dimension, and $\tilde{\mu}_m, \tilde{\sigma}^2_m$ the corresponding statistics averaged from the encoder outputs $\boldsymbol{\mu}_m, \boldsymbol{\sigma}^2_m$:
\begin{align}
  \mathcal{L}_{\mathrm{stat}}
  &= \tfrac{1}{4}\bigl(
    \mathrm{MSE}(\tilde{\mu}_a,\hat{\mu}_a)
    +\mathrm{MSE}(\tilde{\sigma}^2_a,\hat{\sigma}^2_a)
    \notag\\
  &\qquad
    +\mathrm{MSE}(\tilde{\mu}_v,\hat{\mu}_v)
    +\mathrm{MSE}(\tilde{\sigma}^2_v,\hat{\sigma}^2_v)
  \bigr),
\end{align}
where $\mathrm{MSE}(X,Y)=\frac{1}{BL}\lVert X-Y\rVert_2^2$. The feature diversity loss $\mathcal{L}_{\mathrm{div}}$ encourages maximizing information entropy to avoid redundant representations. The overall objective is:
\begin{align}
  \mathcal{L} &=
  \mathcal{L}_{\mathrm{task}}
  + \lambda_{\mathrm{recon}} \mathcal{L}_{\mathrm{recon}}
  + \lambda_{\mathrm{uni}} \mathcal{L}_{\mathrm{uni}}
  \notag\\
  &\quad
  + \lambda_{\mathrm{div}} \mathcal{L}_{\mathrm{div}}
  + \lambda_{\mathrm{stat}} \mathcal{L}_{\mathrm{stat}},
  \label{equation:total_loss}
\end{align}
where $\mathcal{L}_{\mathrm{recon}} = \frac{1}{2}\bigl(\lVert\mathbf{A}-\hat{\mathbf{A}}\rVert_2^2 + \lVert\mathbf{V}-\hat{\mathbf{V}}\rVert_2^2\bigr)$ and $\mathcal{L}_{\mathrm{uni}}$ aggregates the sentiment regression losses from the unimodal heads of each modality.

Step 4: Gradient Modulation with Conflict-aware Penalty (CP).
During training, we balance modality learning via performance-driven gradient modulation. We compute unimodal errors (MAE) and define inverse-error scores; the better-performing modality receives a smaller gradient scale:
\begin{align}
  s_a &= \frac{1}{\mathrm{MAE}_a+\varepsilon}, \qquad
  s_v = \frac{1}{\mathrm{MAE}_v+\varepsilon}, \nonumber \\
  c_a &=
  \begin{cases}
    1-\tanh\left(\alpha\,\operatorname{ReLU}\left(\frac{s_a}{s_v+\varepsilon}\right)\right),
    & s_a>s_v, \\
    1, & \text{otherwise},
  \end{cases}
  \label{equation:gm_coeff}
\end{align}
and $c_v$ is defined symmetrically. To mitigate cases where a modality is already superior but still imposes a disproportionately large gradient norm on shared parameters, the \textbf{Conflict-aware Penalty (CP)} applies an additional multiplicative penalty when this condition is detected:
\begin{align}
  \mathrm{conflict} &= \big(\mathrm{MAE}_a<\mathrm{MAE}_v\big)\ \wedge\ \big(g_a>g_v\big), \nonumber \\
  c_a &\leftarrow c_a \cdot \eta \quad \text{if conflict},
  \label{equation:conflict_penalty}
\end{align}
where $g_a$ and $g_v$ denote the average gradient norms over the corresponding modality encoders, and $\eta\in(0,1)$ is a fixed penalty factor. Crucially, CP keeps gradient pressure balanced across modalities, which also prevents the dominant modality from destabilizing the $\mathcal{L}_{\mathrm{stat}}$ objective---as confirmed by the collapse observed when SL is used without CP (rows A5 and C4 in Table~\ref{tab:ablation}). In practice, modulation is applied only within a scheduled epoch range to stabilize early training while allowing unconstrained convergence later.

\section{Experiments}


\definecolor{best}{RGB}{70,130,180}       
\definecolor{second}{RGB}{173,214,235}    
\definecolor{oursrow}{RGB}{255,242,204}   

\newcommand{\bestcell}[1]{\cellcolor{best}\textbf{#1}}
\newcommand{\secondcell}[1]{\cellcolor{second}\underline{#1}}


\begin{table*}[t]
\centering
\caption{Comparison on CMU-MOSI and CMU-MOSEI. 
\colorbox{best}{Best} and \colorbox{second}{2nd best} results are highlighted.}
\label{tab:mosi_mosei}
\begin{minipage}{0.48\textwidth}
\centering
\small
\textbf{CMU-MOSI}
\begin{tabular}{lcccc}
\toprule
Method & Acc-2$\uparrow$ & F1$\uparrow$ & MAE$\downarrow$ & Corr$\uparrow$ \\
\midrule
Self-MM$_b$~\cite{yu2021learning} & 84.0 & 84.4 & 0.713 & 0.798 \\
MMIM$_b$~\cite{han2021improving} & 84.1 & 84.0 & 0.700 & 0.800 \\
MAG$_b$~\cite{rahman2020integrating} & 84.2 & 84.1 & 0.712 & 0.796 \\
Self-MM$_d$~\cite{yu2021learning} & 55.1 & 53.5 & 1.440 & 0.158 \\
MMIM$_d$~\cite{han2021improving} & 85.8 & 85.9 & 0.649 & 0.829 \\
MAG$_d$~\cite{rahman2020integrating} & 86.1 & 86.0 & 0.690 & 0.831 \\
UniMSE~\cite{hu2022unimse} & 85.9 & 85.8 & 0.691 & 0.809 \\
MIB~\cite{mai2022multimodal} & 85.3 & 85.3 & 0.711 & 0.798 \\
BBFN~\cite{han2021bi} & 84.3 & 84.3 & 0.776 & 0.755 \\
ITHP~\cite{xiao2024neuroinspiredinformationtheoretichierarchicalperception} & \colorbox{second}{88.7} & \colorbox{second}{88.6} & \colorbox{second}{0.643} & \colorbox{second}{0.852} \\
\midrule
\textbf{Ours} & \colorbox{best}{89.31} & \colorbox{best}{89.23} & \colorbox{best}{0.638} & \colorbox{best}{0.864} \\
\bottomrule
\end{tabular}
\end{minipage}
\hfill
\begin{minipage}{0.48\textwidth}
\centering
\small
\textbf{CMU-MOSEI}
\begin{tabular}{lcccc}
\toprule
Method & Acc-2$\uparrow$ & F1$\uparrow$ & MAE$\downarrow$ & Corr$\uparrow$ \\
\midrule
Self-MM$_b$~\cite{yu2021learning} & 85.0 & 85.0 & \colorbox{second}{0.529} & 0.767 \\
MMIM$_b$~\cite{han2021improving} & 86.0 & 86.0 & \colorbox{best}{0.526} & 0.772 \\
MAG$_b$~\cite{rahman2020integrating} & 84.8 & 84.7 & 0.543 & 0.755 \\
Self-MM$_d$~\cite{yu2021learning} & 65.3 & 65.4 & 0.813 & 0.208 \\
MMIM$_d$~\cite{han2021improving} & 85.2 & 85.4 & 0.568 & 0.799 \\
MAG$_d$~\cite{rahman2020integrating} & 85.8 & 85.9 & 0.636 & 0.800 \\
ITHP~\cite{xiao2024neuroinspiredinformationtheoretichierarchicalperception} & \colorbox{second}{87.30} & \colorbox{best}{87.40} & 0.564 & \colorbox{second}{0.813} \\
\midrule
\textbf{Ours} & \colorbox{best}{87.32} & \colorbox{second}{87.22} & 0.552 & \colorbox{best}{0.820} \\
\bottomrule
\end{tabular}
\end{minipage}
\end{table*}

\definecolor{best}{RGB}{70,130,180}
\definecolor{second}{RGB}{173,214,235}
\definecolor{groupA}{RGB}{240,248,255}
\definecolor{groupB}{RGB}{245,245,245}
\definecolor{groupC}{RGB}{255,250,240}
\definecolor{checkgreen}{RGB}{198,239,206}

\newcommand{\best}[1]{\cellcolor{best}\textbf{#1}}
\renewcommand{\secondcell}[1]{\cellcolor{second}\underline{#1}}
\newcommand{\cmark}{\cellcolor{checkgreen}$\checkmark$}

\begin{table}[t]
  \caption{%
    Ablation study results on CMU-MOSI.
    \colorbox{best}{\strut Best} and
    \colorbox{second}{\strut 2nd best} per column are highlighted.%
  }
  \label{tab:ablation}
  \centering
  \setlength{\tabcolsep}{3pt}
  \renewcommand{\arraystretch}{1.12}
  \footnotesize
  \begin{tabular}{c p{2.8cm} cccc}
    \toprule
    \textbf{ID} & \textbf{Configuration}
      & \textbf{Acc-2}$\uparrow$ & \textbf{F1}$\uparrow$
      & \textbf{MAE}$\downarrow$ & \textbf{Corr}$\uparrow$ \\
    \midrule
    A0 & Baseline         & 85.34 & 85.38 & 0.6773 & 82.95 \\
    A1 & +AME             & 86.87 & 86.84 & 0.6771 & 83.14 \\
    A2 & +AME+GM          & 86.41 & 86.39 & 0.6653 & \colorbox{second}{\strut 85.38} \\
    A3 & +AME+GM+GE       & 87.48 & 87.48 & 0.6613 & 85.11 \\
    A4 & +AME+GM+GE+CP    & \colorbox{second}{\strut 87.63} & \colorbox{second}{\strut 87.64} & \colorbox{second}{\strut 0.6560} & 84.78 \\
    A5 & +AME+GM+GE+SL    & 79.24 & 79.32 & 1.1268 & 72.92 \\
    A6 & Full Model (Ours) & \colorbox{best}{\strut 89.31} & \colorbox{best}{\strut 89.23} & \colorbox{best}{\strut 0.6379} & \colorbox{best}{\strut 86.44} \\
    \midrule
    B1 & AME Only         & 86.87 & 86.84 & 0.6771 & 83.14 \\
    B2 & GM Only          & 87.18 & 87.11 & \colorbox{second}{\strut 0.6690} & 83.79 \\
    B3 & SL Only          & \colorbox{second}{\strut 88.09} & \colorbox{second}{\strut 88.07} & \colorbox{best}{\strut 0.6599} & \colorbox{best}{\strut 85.23} \\
    B4 & GM+GE            & \colorbox{best}{\strut 88.24} & \colorbox{best}{\strut 88.22} & 0.6938 & 83.29 \\
    B5 & GM+CP            & 86.87 & 86.83 & 0.6808 & \colorbox{second}{\strut 84.11} \\
    \midrule
    C1 & GM+GE+CP         & 83.05 & 83.11 & 0.8646 & 75.23 \\
    C2 & AME+SL           & 84.89 & 84.96 & 0.7362 & \colorbox{second}{\strut 81.04} \\
    C3 & AME+GM           & \colorbox{second}{\strut 86.41} & \colorbox{second}{\strut 86.39} & \colorbox{second}{\strut 0.6653} & \colorbox{best}{\strut 85.38} \\
    C4 & Full $-$ CP      & 79.24 & 79.32 & 1.1268 & 72.92 \\
    C5 & Full $-$ GE      & \colorbox{best}{\strut 87.48} & \colorbox{best}{\strut 87.37} & \colorbox{best}{\strut 0.6478} & \colorbox{best}{\strut 85.38} \\
    \bottomrule
  \end{tabular}
\end{table}

\begin{figure}[t]
  \centering
  \includegraphics[width=\linewidth]{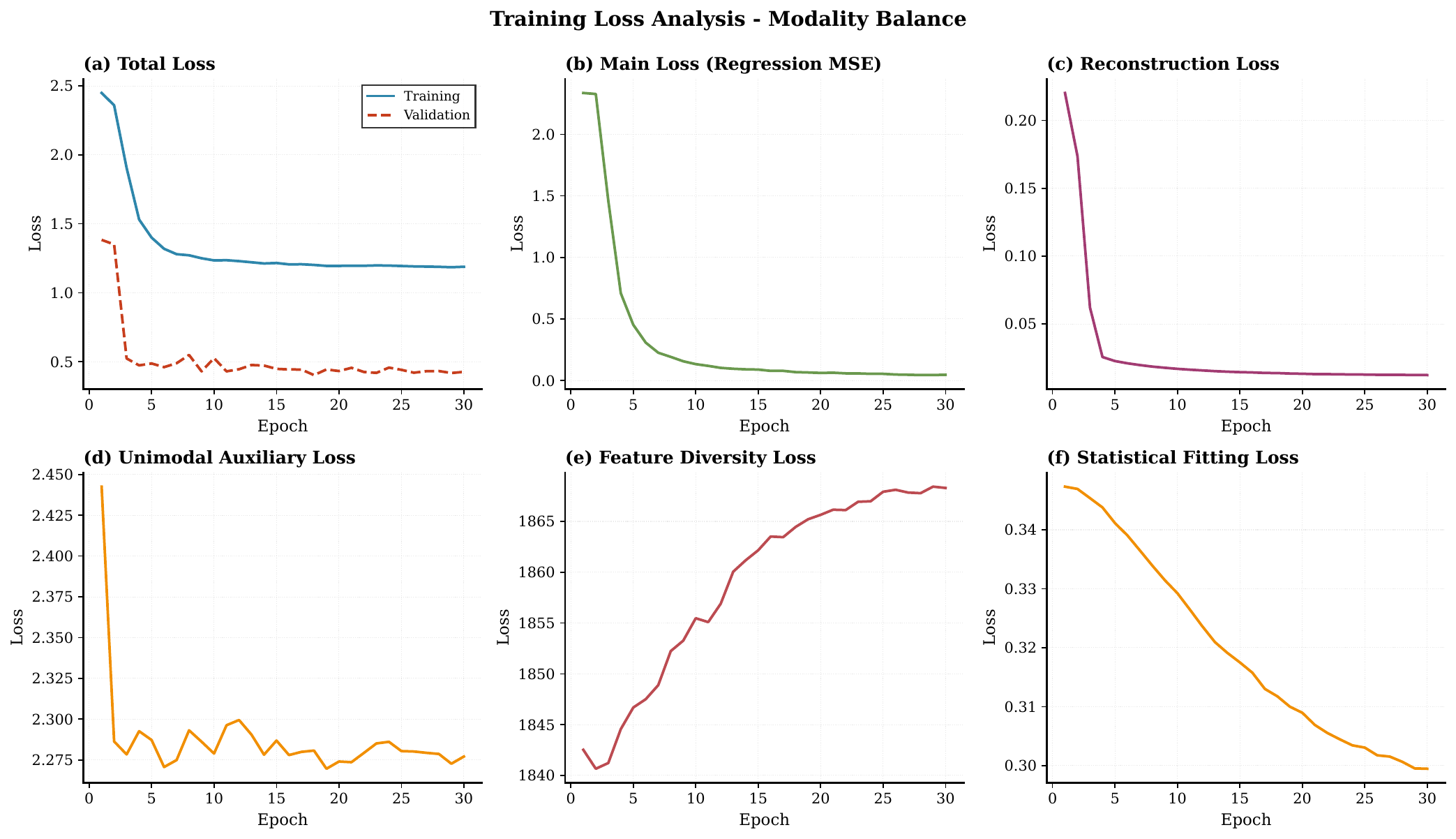}
  \caption{Training loss trajectories across primary and auxiliary objectives on the CMU-MOSI dataset.}
  \label{fig:loss_curves}
\end{figure}

\begin{figure}[t]
  \centering
  \includegraphics[width=\linewidth]{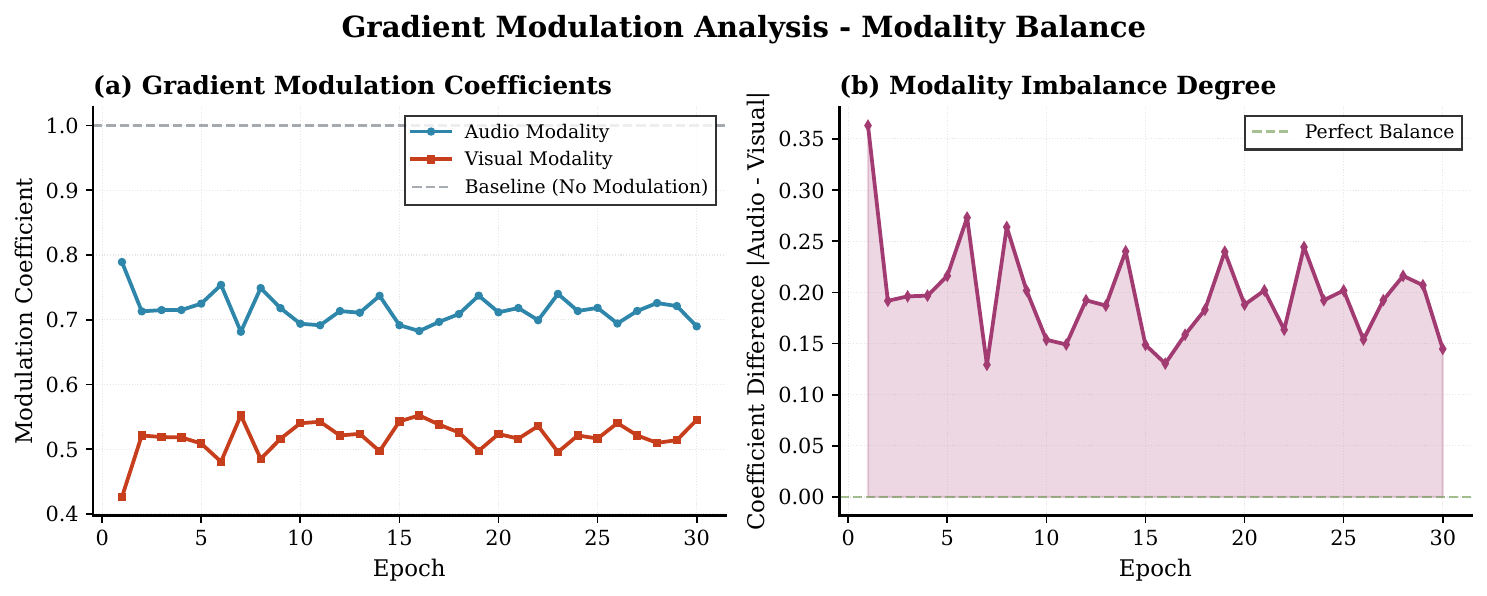}
  \caption{Dynamics of gradient modulation coefficients and the resulting modality imbalance degree.}
  \label{fig:mod_analysis}
\end{figure}

\subsection{Settings}
We evaluate on CMU-MOSI~\cite{zadeh2016mosi} and CMU-MOSEI~\cite{zadeh2018mosei}.
CMU-MOSI comprises 2,199 opinion video clips annotated with sentiment intensity
on a continuous $[-3,3]$ scale, covering aligned text, acoustic, and visual
streams. CMU-MOSEI is a larger benchmark of 23,500 sentence-level utterances
from over 1,000 YouTube speakers across 250 topics, annotated with the same
sentiment scale. We follow the standard train/validation/test splits and report
Acc-2, F1, MAE, and Pearson correlation as in prior work.The text encoder is DeBERTa-v3-base; acoustic and visual streams are aligned
to subword tokens, truncated/padded to length 50, and min--max normalized
per mini-batch. We train for 30 epochs with AdamW (lr$=1\times10^{-5}$,
warmup ratio 0.1) and a training batch size of 8 on 4$\times$ NVIDIA GeForce
RTX 3090 GPUs. The intermediate and fusion dimensions are 256 and 512, with
dropout 0.3 (encoders) and 0.5 (classifier). Loss weights are
$\lambda_{\text{recon}}{=}1.0$, $\lambda_{\text{uni}}{=}0.5$,
$\lambda_{\text{div}}{=}0.1$, $\lambda_{\text{stat}}{=}0.1$, and residual
weight $\beta{=}1.0$. Gradient modulation is applied during epochs 0--25 with
$\alpha{=}1.0$ and conflict penalty factor $\eta{=}0.5$.

\subsection{Results}

Table~\ref{tab:mosi_mosei} presents the comparison between our method and representative baselines on CMU-MOSI and CMU-MOSEI. On CMU-MOSI, our model achieves the best results across all four metrics, reaching 89.31\% binary accuracy, 89.23 F1, 0.638 MAE, and 0.864 Pearson correlation, outperforming the previous best ITHP by a clear margin. On CMU-MOSEI, our model attains the highest Pearson correlation (0.820) and second-best binary accuracy (87.32\%), trailing UniMSE on Acc-2 and F1 while surpassing it on Corr. The consistent improvements on MOSI demonstrate that our adaptive modality encoding and gradient modulation effectively capture complementary cross-modal signals, while the MOSEI results indicate that the proposed framework generalizes well to larger and more diverse speaker conditions. Overall, the results confirm that our method is competitive with or superior to recent state-of-the-art approaches across both benchmarks.

\subsection{Ablation study}
We conduct an ablation study on CMU-MOSI to evaluate the contributions of each component, as shown in Table~\ref{tab:ablation}. 

\textbf{Group A} shows incremental improvements. Adding Adaptive Modality Encoding (AME) improves Acc-2 and F1 (A0→A1). Gradient Modulation (GM) further enhances MAE and Corr (A2), and Gradient Enhancement (GE) improves Acc-2, F1, and MAE (A3), with a marginal decrease in Corr, indicating GE primarily benefits classification-oriented metrics. Including CP improves Acc-2, F1, and MAE (A4), with a trade-off in Corr at this intermediate stage, but Statistical Loss (SL) without CP causes a collapse (A5), as the dominant modality's gradient pressure interferes with distributional regularization. The full model (A6) achieves the best results, demonstrating the need for both CP and SL.

\textbf{Group B} tests each module alone. SL alone (B3) gives strong results, particularly for MAE and Corr. GM+GE (B4) provides the best Acc-2 and F1, while GM+CP (B5) improves Corr, highlighting the complementary roles of GE and CP. Notably, SL alone (B3) performs well precisely because no gradient modulation is active to destabilize it; collapse occurs only when GM is present without CP (A5, C4).

\textbf{Group C} explores critical combinations. Removing CP (C4) causes performance collapse, as seen in A5. Removing GE (C5) leads to minimal degradation, indicating GE's incremental benefit. Critically, GM+GE+CP without AME (C1) severely underperforms (MAE: 0.8646), despite B5 (GM+CP, no AME) achieving MAE: 0.6808. This reveals that GE amplifies gradient signals indiscriminately without AME's stable latent representations; AME is thus a prerequisite for GE to contribute constructively rather than amplify noise. Overall, the full model benefits from synergistic interaction: AME stabilizes features, GM optimizes modalities, and SL+CP stabilize and regularize the training.

\subsection{Analysis of Training Dynamics}

Fig.\ref{fig:loss_curves} shows the loss trajectories over 30 epochs on the MOSI dataset. The Total Loss (a) and Main Regression Loss (b) converge quickly, followed by a stable plateau, indicating stable optimization. The Reconstruction Loss (c) and Statistical Fitting Loss (f) decline smoothly, demonstrating that SL aligns distribution statistics without instability. The Feature Diversity Loss (e) increases, indicating the model maximizes information entropy to avoid redundant representations. The alignment between training and validation loss suggests that CP and SL effectively regularize against overfitting.

\subsection{Gradient Modulation and Modality Balance}

Fig.\ref{fig:mod_analysis} shows the behavior of the Conflict-aware Penalty (CP) and gradient modulation during training. In plot (a), the modulation coefficients for audio and visual modalities stay below the baseline ($1.0$), indicating active gradient suppression to balance their influence. The visual modality receives a lower coefficient (around $0.45$-$0.55$) compared to the audio modality (around $0.7$-$0.8$), suggesting that the CP successfully mitigated the visual modality's excessive gradient pressure. In plot (b), the Modality Imbalance Degree oscillates between $0.15$ and $0.25$, reflecting the dynamic, real-time adjustments that prevent any modality from dominating the optimization process, ensuring a balanced multimodal inference for sentiment prediction.

\section{Conclusion}

In this paper, we address modality imbalance and optimization instability in MSA by introducing a Conflict-aware Penalty (CP) and Statistical Loss (SL). CP mitigates gradient norm conflicts to balance modality optimization, while SL regularizes training by aligning distribution statistics. Experiments on CMU-MOSI demonstrate state-of-the-art performance across all metrics. On the more diverse CMU-MOSEI, the proposed framework achieves the highest Pearson correlation, with remaining metrics competitive with the current best.

\section{Generative AI Use Disclosure}
In the preparation of this manuscript, the authors used generative AI tools (e.g., large language models) solely for the purpose of refining the language and improving the fluency of certain sentences. All research ideas, experimental designs, proposed methods, data analysis, and conclusions were independently conceived and completed by the authors. Generative AI tools were not used to produce any substantive content of this paper.

\bibliographystyle{IEEEtran}
\bibliography{mybib}

\end{document}